\documentclass[runningheads]{llncs}
\usepackage{graphicx}
\usepackage{comment}
\usepackage{amsmath,amssymb} 
\usepackage{color,multirow,wrapfig}

\graphicspath{{figures/}}

\newcommand{\R}{{\mathbb{R}}}
\newcommand{\etal}{{et al.~}}

\begin{document}
\pagestyle{headings}
\mainmatter
\def\ECCVSubNumber{4112}  

\title{Least squares surface reconstruction\\ on arbitrary domains} 

\titlerunning{Least squares surface reconstruction on arbitrary domains}
%
\author{Dizhong Zhu\inst{1}\orcidID{0000-0003-4086-7293} \and
William A. P. Smith\inst{1}\orcidID{0000-0002-6047-0413}}
\authorrunning{Zhu and Smith}
%
\institute{University of York, York, UK \and
\email{\{dizhong.zhu,william.smith\}@york.ac.uk}}
\maketitle

\begin{abstract}
Almost universally in computer vision, when surface derivatives are required, they are computed using only first order accurate finite difference approximations. We propose a new method for computing numerical derivatives based on 2D Savitzky-Golay filters and K-nearest neighbour kernels. The resulting derivative matrices can be used for least squares surface reconstruction over arbitrary (even disconnected) domains in the presence of large noise and allowing for higher order polynomial local surface approximations. They are useful for a range of tasks including normal-from-depth (i.e.~surface differentiation), height-from-normals (i.e.~surface integration) and shape-from-x. We show how to write both orthographic or perspective height-from-normals as a linear least squares problem using the same formulation and avoiding a nonlinear change of variables in the perspective case. We demonstrate improved performance relative to state-of-the-art across these tasks on both synthetic and real data and make available an open source implementation of our method.
\end{abstract}

\section{Introduction}
\label{sec:intro}

Estimating derivatives of a noisy measured signal is a basic problem in signal processing and finds application in areas ranging from spectroscopy to finance. The inverse of this problem arises when reconstructing a function from noisy measurements of its derivatives. This is a common problem in computer vision when estimating a surface (either an orthographic relative height map or a perspective absolute depth map) from noisy measurements of the surface normals or 2D surface gradient. This problem is usually known as surface integration or height-from-gradient \cite{queau2018normal}. More generally, shape-from-x methods that use a surface orientation cue to directly reconstruct a discrete 3D surface representation also require numerical approximations of the surface derivative. These approaches include shape-from-shading \cite{durou2008numerical}, polarisation \cite{zhu2019depth}, texture \cite{clerc2002texture} and photometric stereo \cite{ackermann2015survey}. In addition, merging depth and surface normal estimates \cite{nehab2005efficiently} requires a derivative operator to relate the two. Finally, recent work on deep depth estimation computes surface gradients in-network so that either surface gradient supervision can be used \cite{li2018megadepth} or to compute surface normals from depth maps \cite{eigen2015predicting}. Hence, numerical surface derivative approximations are of fundamental importance in computer vision. 

It is therefore surprising that almost universally in the surface reconstruction literature, numerical derivative approximations that are only first order accurate (forward or backward finite difference) are used that make an implicit assumption of surface planarity and are highly susceptible to noise. Occasionally, central difference (second order accurate) \cite{queau2018variational} or smoothed central difference (increased robustness to noise) \cite{nehab2005efficiently} kernels have been used but the only work to consider kernels accurate to arbitrary order is that of Harker and O'Leary \cite{harker2008least,harker2011least,harker2013direct,harker2015regularized}. However, their formulation works only on a rectangular domain meaning it cannot be applied to objects with arbitrary foreground masks. In addition, they use 1D kernels which cannot gain robustness by using a local neighbourhood spanning different rows and columns.

In this paper, we extend the idea of least squares surface integration in a number of ways. Like Harker and O'Leary we use kernels that are higher order accurate but, differently, we allow for arbitrary, even disconnected, domains. To the best of our knowledge, we are the first to use 2D Savitzky-Golay filters over an arbitrary domain for surface reconstruction problems (height-from-gradient and shape-from-x). Second, we propose to also use Savitzky-Golay filters as a smoothness regulariser. Unlike planar regularisers, such as a Laplacian filter \cite{smith2019height} or zero surface prior \cite{queau2018variational}, we are able to use a high regularisation weight to cope with very significant noise, yet still recover smooth curved surfaces without over flattening. Third, our least squares surface reconstruction approach is very general, allowing both orthographic and perspective projection (without requiring a nonlinear change of variables \cite{queau2018normal}), and an optional depth prior. Fourth, we propose an alternate formulation for height-from-normals that uses surface normal components rather than implied surface gradients and is numerically more stable. Finally, we make available an open source implementation of the methods that can easily be integrated into a surface reconstruction pipeline.

\subsection{Related work}

Computing differential surface properties from discrete surface representations is a large topic within computer graphics. Of particular interest in this work is the task of computing surface normals from potentially noisy depth maps. Mitra \etal \cite{mitra2003estimating} describe a classical approach in which surfaces are locally approximated by a plane fitted by least squares to nearest neighbour points. Klassing \etal \cite{klasing2009comparison} compare a variety of approaches and conclude that the straightforward plane PCA method performs well. Comino \etal \cite{comino2018sensor} incorporate knowledge of the sensing device in order to develop an adaptive algorithm.

Classical approaches to recovering surface height from the surface normal or gradient field are based on the line integral \cite{klette1998three,robles2005graph,wu1988line}. They optimise local least squares cost functions and differ in their path selection strategy. Global methods were pioneered by Horn and Brooks \cite{horn1986variational}, who posed the problem in the continuous domain as a least squares optimisation problem. Although not convergent or practical, this approach led the way to many more modern approaches. Frankot and Chellappa \cite{frankot1988method} solved the same problem but formulated on the Fourier basis with a fast algorithm based on the DFT. Kovesi \cite{kovesi2005shapelets} uses a shapelet basis instead. Both methods assume periodic boundary conditions that introduces bias.

Agrawal \etal \cite{agrawal2006range} construct a discrete Poisson equation and solve it efficiently. However, they use forward/backward finite difference approximations and a zero gradient boundary assumptions that biases the reconstruction. Simchony \etal \cite{simchony1990direct} solve a Poisson equation using the Discrete Cosine Transform but require a rectangular domain. Harker and O'Leary \cite{harker2008least,harker2011least,harker2013direct,harker2015regularized} proposed a least squares approach and a subsequent series of refinements including a variety of regularisers. Their formulation is based on a rectangular matrix representation for the unknown height field. In this case, the numerical derivatives can be obtained by pre and post multiplication with an appropriate derivative matrix $\mathbf{D}_v\mathbf{Z}$ and $\mathbf{Z}\mathbf{D}_u$ where surface heights are stored in a matrix the same size as the image, $\mathbf{Z}\in\R^{W\times H}$. The differentiation matrices are both square with size equal to the height and width of the image respectively, $\mathbf{D}_u\in\R^{H\times H}$ and $\mathbf{D}_v\in\R^{W\times W}$. Hence, their size is $O(n)$ for $n=WH$ pixels. Harker and O'Leary show that the least squares problem can be written as a Sylvester equation and solved extremely efficiently. For robustness in the presence of noise, it is important to use local context to assist in the computation of the derivatives. The drawback of these $O(n)$ derivative matrices is that they can only use a neighbourhood of pixels in the same row (for horizontal derivatives) or column (for vertical). But the most significant drawback of their approach is the requirement for a rectangular domain. This rarely holds when either dealing with objects with a foreground mask or noisy sensor data with holes.

Recently, Qu{\'e}au \etal \cite{queau2018variational} proposed the state-of-the-art method based on solving a least squares system formulated using sparse differentiation matrices $\mathbf{D}_u\mathbf{z}$ and $\mathbf{D}_v\mathbf{z}$ where foreground surface heights are stored in a vector with arbitrary ordering, $\mathbf{z}\in\R^{n}$. The differentiation matrices are both of the same size, $\mathbf{D}_u,\mathbf{D}_v\in\R^{n\times n}$. In the case where all pixels are foreground, $n=HW$ and these matrices are very large. However, they are sparse since each row has non-zero values only in columns corresponding to pixels in the local region of the pixel being differentiated. In the minimal case (forward, backward or central difference), each row has only two non-zero values. In contrast to Harker and O'Leary, this approach can deal with arbitrary domains. However, unlike Harker and O'Leary, it uses an average of forward and backward finite difference (central difference) which are not exact for higher order surfaces.

All of these methods make the assumption of orthographic projection. Qu{\'e}au \etal \cite{queau2018normal} point out that any orthographic algorithm can be used for perspective surface integration by a nonlinear change of variables by solving in the log domain. The drawback of this transformation is that the solution is only least squares optimal in the transformed domain. When exponentiating to recover the perspective surface, large spikes can occur. The only method formulated in the perspective domain that we are aware of is that of Nehab \etal \cite{nehab2005efficiently}, though in a slightly different context of merging depth and normals. Moreover, they use a derivative approximation based on smoothed central difference.

In this paper we bring together the best of both of these formulations and propose an approach that can handle arbitrary domains and uses arbitrary order numerical derivative approximations. Moreover, we reformulate the least squares height from normals problem such that it can handle both perspective and orthographic projection models.

\section{Linear least squares height-from-normals}

We denote a 3D point in world units as $\mathbf{p}=(x,y,z)$ and an image location in camera units (pixels) as $(u,v)$ such that $\mathbf{u}=(u,v)$ is a pixel location in the image. We parameterise the surface by the height or depth function $z(\mathbf{u})$. In normals-from-depth we are given a noisy observed depth map and wish to estimate the surface normal map $\mathbf{n}(\mathbf{u})=[n_x(\mathbf{u}),n_y(\mathbf{u}),n_z(\mathbf{u})]^T$ with $\|\mathbf{n}(\mathbf{u})\|=1$. In surface integration we are given $\mathbf{n}(\mathbf{u})$ and wish to estimate $z(\mathbf{u})$. 

To the best of our knowledge, all existing methods compute height-from-\emph{gradient}, i.e.~they transform the given surface normals into the surface gradient and solve the following pair of PDEs, usually in a least squares sense:
\begin{equation}
    \frac{\partial z(\mathbf{u})}{\partial u} = \frac{-n_x(\mathbf{u})}{n_z(\mathbf{u})},\quad
    \frac{\partial z(\mathbf{u})}{\partial v} = \frac{-n_y(\mathbf{u})}{n_z(\mathbf{u})}.\label{eqn:classical}
\end{equation}
The problem with this approach is that close to the occluding boundary, $n_z$ gets very small making the gradient very large. The squared errors in these pixels then dominate the least squares solution. We propose an alternative formulation that is more natural, works with both orthographic and perspective projections and, since it uses the components of the normals directly, is best referred to as \emph{height-from-normals}. The idea is that the surface normal should be perpendicular to the tangent vectors. This leads to a pair of PDEs:
\begin{equation}
    \frac{\partial \mathbf{p}(\mathbf{u})}{\partial u} \cdot \mathbf{n}(\mathbf{u}) = 0,\quad
    \frac{\partial \mathbf{p}(\mathbf{u})}{\partial v} \cdot \mathbf{n}(\mathbf{u}) = 0,\label{eqn:tangentconst}
\end{equation}
where $\mathbf{p}(\mathbf{u})$ denotes the 3D position corresponding to pixel position $\mathbf{u}$ and $\frac{\partial \mathbf{p}(\mathbf{u})}{\partial u}$, $\frac{\partial \mathbf{p}(\mathbf{u})}{\partial v}$ are the image plane derivatives (i.e.~partial derivatives with respect to pixel coordinates) of the 3D point position. We now consider how to formulate equations of this form in two different cases: orthographic and perspective projection.

\subsection{Linear equations}

\paragraph{Orthographic case} The 3D position, $\mathbf{p}(\mathbf{u})$, of the point on the surface that projects to pixel position $\mathbf{u}$ and its derivatives are given by:
\begin{equation} 
    \mathbf{p}(\mathbf{u}) = 
    \begin{bmatrix} 
    u\\
    v\\
    z(\mathbf{u})
    \end{bmatrix},\quad
    \frac{\partial \mathbf{p}(\mathbf{u})}{\partial u} = 
    \begin{bmatrix} 
    1\\
    0\\
    \frac{\partial z(\mathbf{u})}{\partial u}
    \end{bmatrix},\quad
    \frac{\partial \mathbf{p}(\mathbf{u})}{\partial v} = 
    \begin{bmatrix} 
    0\\
    1\\
    \frac{\partial z(\mathbf{u})}{\partial v}
    \end{bmatrix}.
\end{equation}
Substituting these derivatives into \eqref{eqn:tangentconst} we obtain:
\begin{equation}
  \frac{\partial z(\mathbf{u})}{\partial u} n_z(\mathbf{u}) = -n_x(\mathbf{u}),\quad
  \frac{\partial z(\mathbf{u})}{\partial v} n_z(\mathbf{u}) = -n_y(\mathbf{u}).\label{eqn:ortholin}
\end{equation}
Note that this is a simple rearrangement of \eqref{eqn:classical} but which avoids division by $n_z$.

\paragraph{Perspective case} In the perspective case, the 3D coordinate corresponding to the surface point at $\mathbf{u}$ and its derivatives are given by:
\begin{equation} 
    \mathbf{p}(\mathbf{u}) = 
    \begin{bmatrix} 
    \frac{u-c_u}{f}z(\mathbf{u})\\
    \frac{v-c_v}{f}z(\mathbf{u})\\
    z(\mathbf{u})
    \end{bmatrix},\label{eqn:pesrppos}
\end{equation}
where $f$ is the focal length of the camera and $(c_u,c_v)$ is the principal point. The derivatives are given by:
\begin{equation}
    \frac{\partial \mathbf{p}(\mathbf{u})}{\partial u}\!=\!\begin{bmatrix} 
    \frac{1}{f}\left((u-c_u)\frac{\partial z(\mathbf{u})}{\partial u}+z(\mathbf{u})\right)\\ 
    \frac{1}{f}(v-c_v)\frac{\partial z(\mathbf{u})}{\partial u}\\ 
    \frac{\partial z(\mathbf{u})}{\partial u}
    \end{bmatrix}\!,\ \ 
    \frac{\partial \mathbf{p}(\mathbf{u})}{\partial v}\!=\!\begin{bmatrix} 
    \frac{1}{f}(u-c_u)\frac{\partial z(\mathbf{u})}{\partial v}\\ 
    \frac{1}{f}\left((v-c_v)\frac{\partial z(\mathbf{u})}{\partial v}+z(\mathbf{u})\right)\\ 
    \frac{\partial z(\mathbf{u})}{\partial v}
    \end{bmatrix}\!.\label{eqn:perspderivs}
\end{equation}
Again, these can be substituted into \eqref{eqn:tangentconst} to relate the derivatives of $z$ to the surface normal direction.

\subsection{Discrete formulation}

Assume that we are given a foreground mask comprising some subset of the discretised image domain, $\mathcal{F}\subseteq \{1,\dots,W\}\times\{1,\dots,H\}$ with $|\mathcal{F}|=n$. The depth values for the $n$ foreground pixels are stored in a vector $\mathbf{z}\in\R^{n}$ with arbitrary ordering. We make use of a pair of matrices, $\mathbf{D}_{\text{u}},\mathbf{D}_{\text{v}}\in\R^{n\times n}$, that compute discrete approximations to the partial derivative in the horizontal and vertical directions respectively. The exact form of these matrices is discussed in the next section. Once these discrete approximations are used, the PDEs in \eqref{eqn:tangentconst} become linear systems of equations in $\mathbf{z}$. This leads to a linear least squares formulation for the height-from-normals problem.

\paragraph{Orthographic case} In the orthographic case, we stack equations of the form \eqref{eqn:ortholin}:
\begin{equation}
    \begin{bmatrix}
    \text{diag}(\mathbf{n}_z)\mathbf{D}_{\text{u}}\\
    \text{diag}(\mathbf{n}_z)\mathbf{D}_{\text{v}}
    \end{bmatrix}\mathbf{z} = 
    \begin{bmatrix}
    -\mathbf{n}_{\text{x}}\\
    -\mathbf{n}_{\text{y}}\\
    \end{bmatrix}\label{eqn:ortholinsystem}
\end{equation}
where 
\begin{equation}
    \mathbf{n}_{\text{x}}=
    \begin{bmatrix}
    n_x(\mathbf{u}_1)\\
    \vdots\\
    n_x(\mathbf{u}_n)
    \end{bmatrix},\quad
    \mathbf{n}_{\text{y}}=
    \begin{bmatrix}
    n_y(\mathbf{u}_1)\\
    \vdots\\
    n_y(\mathbf{u}_n)
    \end{bmatrix},\quad
    \mathbf{n}_{\text{z}}=
    \begin{bmatrix}
    n_z(\mathbf{u}_1)\\
    \vdots\\
    n_z(\mathbf{u}_n)
    \end{bmatrix}.
\end{equation}
Note that \eqref{eqn:ortholinsystem} is satisfied by any offset of the true $\mathbf{z}$, corresponding to the unknown constant of integration. This is reflected in the fact that:
\begin{equation}
 \text{rank}\left(\begin{bmatrix}\mathbf{D}_{\text{u}}\\\mathbf{D}_{\text{v}}\end{bmatrix}\right)=n-1.
\end{equation}
So, in the orthographic case, we can only recover $\mathbf{z}$ up to an unknown offset.

\paragraph{Perspective case} In the perspective case, we stack equations obtained by substituting \eqref{eqn:perspderivs} in \eqref{eqn:tangentconst} to obtain:
\begin{equation}
    \begin{bmatrix}
    \mathbf{NT}_x \\
    \mathbf{NT}_y \\
    \end{bmatrix}\mathbf{z}=
    \mathbf{0}_{2n\times 1},\label{eqn:persplinsystem}
\end{equation}
where
\begin{equation}
    \mathbf{T}_x = \begin{bmatrix}
    \frac{1}{f}\mathbf{U} & \frac{1}{f}\mathbf{I} \\
    \frac{1}{f}\mathbf{V} & \mathbf{0}_{n\times n} \\
    \mathbf{I} & \mathbf{0}_{n\times n} \\
    \end{bmatrix}
    \begin{bmatrix}\mathbf{D}_u \\ \mathbf{I} 
    \end{bmatrix},\ \  
    \mathbf{T}_y = \begin{bmatrix}
    \frac{1}{f}\mathbf{U} & \mathbf{0}_{n\times n} \\
    \frac{1}{f}\mathbf{V} & \frac{1}{f}\mathbf{I} \\
    \mathbf{I} & \mathbf{0}_{n\times n} \\
    \end{bmatrix}
    \begin{bmatrix}\mathbf{D}_v \\ \mathbf{I} 
    \end{bmatrix},\ \ 
    \mathbf{N}=\begin{bmatrix}
    \text{diag}\left( \mathbf{n}_{\text{x}} \right) \\
    \text{diag}\left( \mathbf{n}_{\text{y}} \right) \\
    \text{diag}\left( \mathbf{n}_{\text{z}} \right)
    \end{bmatrix}^T\!\!,
\end{equation}
$\mathbf{U}=\textrm{diag}(u_1-c_u,\dots,u_n-c_u)$ and $\mathbf{V}=\textrm{diag}(v_1-c_v,\dots,v_n-c_v)$. Note that \eqref{eqn:persplinsystem} is a homogeneous linear system. This means that it is also satisfied by any scaling of the true $\mathbf{z}$. So, in the perspective case, we can only recover $\mathbf{z}$ up to an unknown scaling.

\section{Numerical differentiation kernels}

We now consider the precise form of $\mathbf{D}_{\text{u}}$ and $\mathbf{D}_{\text{v}}$ and propose a novel alternative with attractive properties. Since the derivative matrices act linearly on $\mathbf{z}$ they can be viewed as 2D convolutions over $z(u,v)$. Note however that each row of $\mathbf{D}_{\text{u}}$ or $\mathbf{D}_{\text{v}}$ can be different - i.e.~different convolution kernels can be used at different spatial locations.

By far the most commonly used numerical differentiation kernels are forward (fw) and backward (bw) difference, shown here for both the horizontal (h) and vertical (v) directions:
\scriptsize
\begin{equation}
\mathbf{K}^{\text{h}}_{\text{fw}}=
    \begin{bmatrix}
    0 & 0 & 0 \\
    0 & -1 & 1 \\
    0 & 0 & 0
    \end{bmatrix},\quad
\mathbf{K}^{\text{v}}_{\text{fw}}=
    \begin{bmatrix}
    0 & 0 & 0 \\
    0 & -1 & 0 \\
    0 & 1 & 0
    \end{bmatrix},\quad
\mathbf{K}^{\text{h}}_{\text{bw}}=
    \begin{bmatrix}
    0 & 0 & 0 \\
    -1 & 1 & 0 \\
    0 & 0 & 0
    \end{bmatrix},\quad
\mathbf{K}^{\text{v}}_{\text{bw}}=
    \begin{bmatrix}
    0 & -1 & 0 \\
    0 & 1 & 0 \\
    0 & 0 & 0
    \end{bmatrix}.
\end{equation}
\normalsize
As resolution increases and the effective step size decreases, forward and backward differences tend towards the exact derivatives. However, for finite step size they are only exact for order one (planar) surfaces and highly sensitive to noise. Averaging forward and backward yields the central difference (c) approximation, used for example by Qu{\'e}au \etal \cite{queau2018variational}:
\scriptsize
\begin{equation}
\mathbf{K}^{\text{h}}_{\text{c}}=
    \frac{1}{2}\begin{bmatrix}
    0 & 0 & 0 \\
    -1 & 0 & 1 \\
    0 & 0 & 0
    \end{bmatrix},\quad
\mathbf{K}^{\text{v}}_{\text{c}}=
    \frac{1}{2}\begin{bmatrix}
    0 & -1 & 0 \\
    0 & 0 & 0 \\
    0 & 1 & 0
    \end{bmatrix}.
\end{equation}
\normalsize
This is order two accurate but still only uses two pixels per derivative and so is sensitive to noise. One way to address this is to first smooth the $z$ values with a smoothing kernel $\mathbf{S}$ and then compute a finite difference approximation. By associativity of the convolution operator we can combine the smoothing and finite difference kernels into a single kernel. For example, the smoothed central difference (sc) approximation, as used by Nehab \etal \cite{nehab2005efficiently} is given by:
\scriptsize
\begin{equation}
\mathbf{K}^{\text{h}}_{\text{sc}}=\mathbf{K}^{\text{h}}_{\text{c}}\ast \mathbf{S}=
    \frac{1}{12}\begin{bmatrix}
    -1 & 0 & 1 \\
    -4 & 0 & 4 \\
    -1 & 0 & 1
    \end{bmatrix},\quad
    \mathbf{K}^{\text{v}}_{\text{sc}}=\mathbf{K}^{\text{v}}_{\text{c}}\ast \mathbf{S}=
    \frac{1}{12}\begin{bmatrix}
    -1 & -4 & -1 \\
    0 & 0 & 0 \\
    1 & 4 & 1
    \end{bmatrix},
\end{equation}
\normalsize
where in this case $\mathbf{S}$ is a rounded approximation of a $3\times 3$ Gaussian filter with standard deviation 0.6. A problem with both smoothed and unsmoothed central difference is that the derivatives and therefore the linear equations for a given pixel do not depend on the height of that pixel. This lack of dependence between adjacent pixels causes a severe ``checkerboard'' effect that necessitates the use of an additional regulariser, often smoothness. Commonly, this is the discrete Laplacian \cite{smith2019height}. However, a smoothness penalty based on this filter is minimised by a planar surface. So, as the regularisation weight is increased, the surface becomes increasingly flattened until it approaches a plane.

With all of these methods alternative kernels must be used at the boundary of the foreground domain. For example, switching from central to backward differences. This means that the numerical derivatives are not based on a consistent assumption. 

\subsection{2D Savitzky-Golay filters}

We now show how to overcome the limitations of the common numerical differentiation and smoothing kernels using 2D Savitzky-Golay filters.

The idea of a Savitzky-Golay filter \cite{savitzky1964smoothing} is to approximate a function in a local neighbourhood by a polynomial of chosen order. This polynomial is fitted to the observed (noisy) function values in the local neighbourhood by linear least squares. Although the polynomial may be of arbitrarily high order, the fit residuals are linear in the polynomial coefficients and so a closed form solution can be found. This solution depends only on the relative coordinates of the pixels in the local neighbourhood. So, it can be applied (linearly) to any data values meaning that reconstruction with the arbitrary order polynomial can be accomplished with a straightforward (linear) convolution.

The surface around a point $(u_0,v_0)$ is approximated by the order $k$ polynomial $z_{u_0,v_0}(u,v):\R^2\mapsto\R$ with coefficients $a_{ij}$:
\begin{equation}
    z_{u_0,v_0}(u,v)=\sum_{i=0}^k\sum_{j=0}^{k-i} a_{ij}(u-u_0)^i(v-v_0)^j.
\end{equation}
Assume we are given a set of pixel locations, $\mathcal{N}_{u_0,v_0}=\{(u_1,v_1),\dots,(u_m,v_m)\}$, forming a neighbourhood around $(u_0,v_0)$ and the corresponding $z$ values for those pixels. We can form a set of linear equations
\begin{equation}\begin{bmatrix}
1,&v_1-v_0,&(v_1-v_0)^2,&\dots,&(u_1-u_0)^k\\
& & & \vdots & \\
1,&v_m-v_0,&(v_m-v_0)^2,&\dots,&(u_m-u_0)^k\\
\end{bmatrix}
\mathbf{a}=\mathbf{C}_{\mathcal{N}_{u_0,v_0}}\mathbf{a}
= \mathbf{z}_{\mathcal{N}_{u_0,v_0}},
\end{equation}
where $\mathbf{a}=[a_{00},a_{01},a_{02},\dots,a_{k0}]^T$ and $\mathbf{z}_{\mathcal{N}_{u_0,v_0}}=[z(u_1,v_1),\dots,z(u_m,v_m)]^T$. The least squares solution for $\mathbf{a}$ is given by $\mathbf{C}^+_{\mathcal{N}_{u_0,v_0}}\mathbf{z}_{\mathcal{N}_{u_0,v_0}}$ where $\mathbf{C}^+_{\mathcal{N}_{u_0,v_0}}$ is the pseudoinverse of $\mathbf{C}_{\mathcal{N}_{u_0,v_0}}$. Note that $\mathbf{C}^+_{\mathcal{N}_{u_0,v_0}}$ depends only on the relative coordinates of the pixels chosen to lie in the neighbourhood of the $(u_0,v_0)$. Also note that $z_{u_0,v_0}(0,0)$ is given simply by $a_{00}$ which is the convolution between the first row of $\mathbf{C}^+_{\mathcal{N}_{u_0,v_0}}$ and the $z$ values. This is a smoothed version of $z(u_0,v_0)$ in which the original surface is locally approximated by a best fit, order $k$ polynomial. Similarly, the first derivative of the fitted polynomial in the horizontal direction is given by $a_{10}$ and in the vertical direction by $a_{01}$, corresponding to two other rows of $\mathbf{C}^+_{\mathcal{N}_{u_0,v_0}}$. Note that the order $k$ is limited by the size of the neighbourhood. Specifically, we require at least as many pixels as coefficients, i.e.~$k\leq m$.

When $\mathcal{N}_{u_0,v_0}$ is a square neighbourhood centred on $(u_0,v_0)$ then the appropriate row of $\mathbf{C}^+_{\mathcal{N}_{u_0,v_0}}$ can be reshaped into a square convolution kernel. Convolving this with a $z(u,v)$ map with rectangular domain $\mathcal{F}$ amounts to locally fitting a polynomial of order $k$ and either evaluating the polynomial at the central position, acting as a smoothing kernel, or evaluating the derivative of the polynomial in either vertical or horizontal direction.

\subsection{K-nearest pixels kernel}

\begin{figure}[!t]
    \centering
    \includegraphics[width=0.9\textwidth,clip=true]{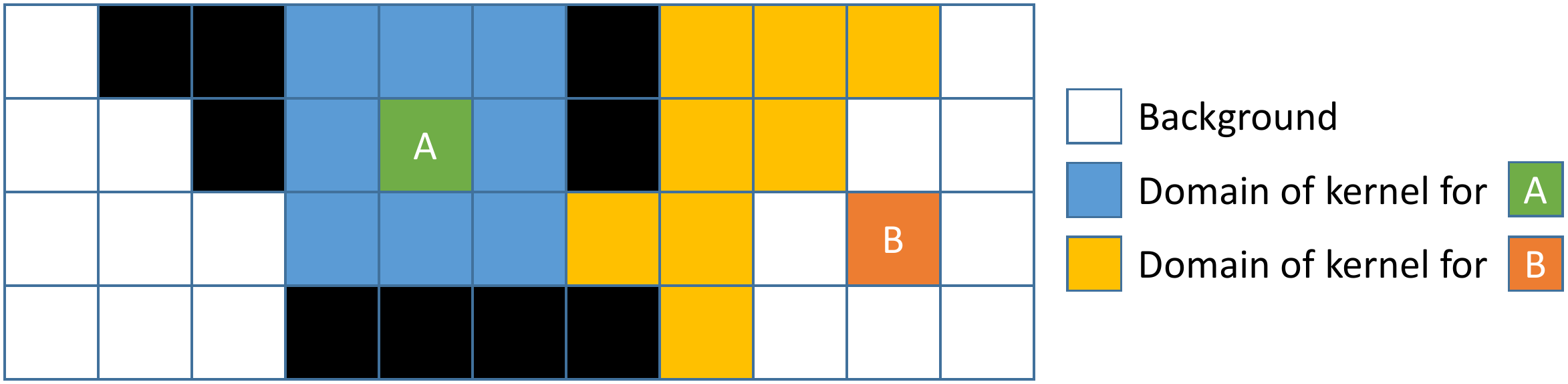}
\scriptsize
\begin{equation*}
\mathbf{K}_{\text{A}}=
\begin{bmatrix}
0 & 0 & 0 & 0 & 0 & 0 & 0 \\
0 & 0 & 0 & 0 & 0 & 0 & 0 \\
0 & 0 & -0.167 & 0 & 0.167 & 0 & 0 \\
0 & 0 & -0.167 & 0 & 0.167 & 0 & 0 \\
0 & 0 & -0.167 & 0 & 0.167 & 0 & 0 \\
0 & 0 & 0 & 0 & 0 & 0 & 0 \\
0 & 0 & 0 & 0 & 0 & 0 & 0
\end{bmatrix},\quad
\mathbf{K}_{\text{B}}=
\begin{bmatrix}
0 & 0 & 0 & 0 & 0 & 0 & 0 \\
0 & 0.266 &  -0.255  &  0.249 & 0 & 0 & 0 \\
0 & -0.294 &  -0.487 & 0 & 0 & 0 & 0 \\
0.4194 &  -0.4698   &      0  &  0.8313 & 0 & 0 & 0 \\
0 & -0.2603 & 0 & 0 & 0 & 0 & 0 \\
0 & 0 & 0 & 0 & 0 & 0 & 0 \\
0 & 0 & 0 & 0 & 0 & 0 & 0
\end{bmatrix}
\end{equation*}
\normalsize
\caption{An example of computing 2D Savitzky-Golay filters on an arbitrary domain. In this example, we use a $3\times 3$ kernel. For point $A$ we can use the default square kernel. The order two Savitzky-Golay filter for the horizontal derivative is shown below as $\mathbf{K}_A$. For point $B$ we use the $3^2$ nearest pixels and build a custom order two Savitzky-Golay filter shown below as $\mathbf{K}_B$. In practice, higher order kernels provide better performance.}
    \label{fig:domain}
\end{figure}

In general, the foreground domain will not be rectangular. Often, it corresponds to an object mask or semantic segmentation of a scene. In this case, we need a strategy to deal with pixels that do not have the neighbours required to use the square kernel. 2D Savitzky-Golay filters are ideal for this because the method described above for constructing them can be used for arbitrary local neighbourhoods. We propose to use the K-nearest pixels in $\mathcal{F}$ to a given pixel. In practice, we compute the square $d\times d$ kernel once and use this for all pixels where the required neighbours lie in $\mathcal{F}$. For those that do not, we find the $d^2$ nearest neighbours in $\mathcal{F}$ (one of which will be the pixel itself). Where tie-breaks are needed, we do so randomly, though we observed no significant difference in performance if all tied pixels are included. In Figure \ref{fig:domain} we show an example of a standard and non-standard case. All non-white pixels lie in $\mathcal{F}$. Pixel $A$ has the available neighbours to use the square kernel while $B$ does not and uses a custom kernel. 

Each element in a kernel for a pixel is copied to the appropriate entries in a row of $\mathbf{D}_{\text{u}}$ or $\mathbf{D}_{\text{v}}$. We similarly construct a matrix $\mathbf{S}\in\R^{n\times n}$ containing the $a_{00}$ kernels, i.e.~the smoothing kernel. Each row of these three matrices has $d^2$ non-zero entries.

\subsection{3D K-nearest neighbours kernel}

For normals-from-depth where a noisy depth map is provided, the K-nearest neighbours kernel idea can be extended to 3D. The idea is to use the depth map with \eqref{eqn:pesrppos} to transform pixels to 3D locations, then to perform the KNN search in 3D. The advantage of this is that kernels will avoid sampling across depth discontinuities where the large change in depth will result in adjacent pixels being far apart in 3D distance. This allows us to create large, robust kernels but without smoothing over depth discontinuities.

\section{Implementation}

For an efficient implementation, all pixel coordinates from $\mathcal{F}$ are placed in a KNN search tree so that local neighbourhoods can be found quickly and pixels that can use the square mask are identified by convolution of the mask with a square filter of ones.

To compute normals-from-depth, we use our proposed derivative matrices (with 3D KNN search) to compute the partial derivatives of $z$, take the cross product between horizontal and vertical derivatives \eqref{eqn:perspderivs} and normalise to give the unit surface normal.

To compute height-from-normals, we solve a system of the form of \eqref{eqn:ortholinsystem} (orthographic) or \eqref{eqn:persplinsystem} (perspective). We augment the system of equations with a smoothness penalty of the form $\lambda(\mathbf{S}-\mathbf{I})\mathbf{z}=\mathbf{0}$, where $\lambda$ is the regularisation weight. This encourages the difference between the smoothed and reconstructed $z$ values to be zero. For the orthographic system, we resolve the unknown offset by solving for the minimum norm solution - equivalent to forcing the mean $z$ value to zero. For the perspective case, since the system is homogeneous in theory we could solve for the $\|\mathbf{z}\|=1$ solution by solving a minimum direction problem using the sparse SVD. In practice, we find it is faster to add an additional equation forcing the solution at one pixel to unity. Finally, we can optionally include a depth prior simply by adding the linear equation $\omega\mathbf{I}\mathbf{z}=\omega\mathbf{z}_{\text{prior}}$, where $\omega$ is the prior weight.

We provide a complete implementation of our method in Matlab\footnote{\url{https://github.com/waps101/LSQSurfaceReconstruction}}.

\section{Evaluation}

\begin{figure}[!t]
    \centering
    \scriptsize{
    \begin{tabular}{ccccc}
        Depth GT & Normals GT & Ours & Finite difference & PCA \cite{klasing2009comparison} \\
        \includegraphics[width=2.3cm]{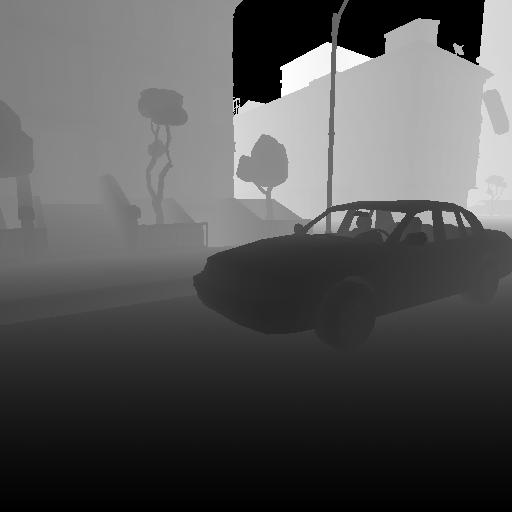} &
        \includegraphics[width=2.3cm]{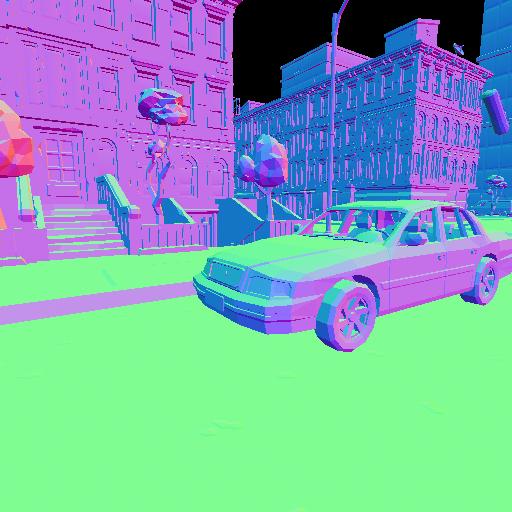} &
        \includegraphics[width=2.3cm]{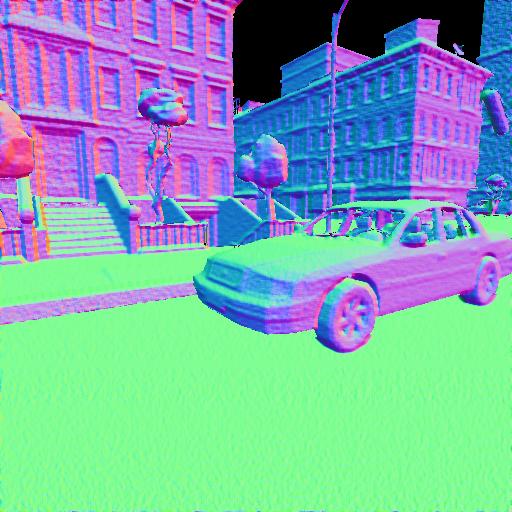} & 
        \includegraphics[width=2.3cm]{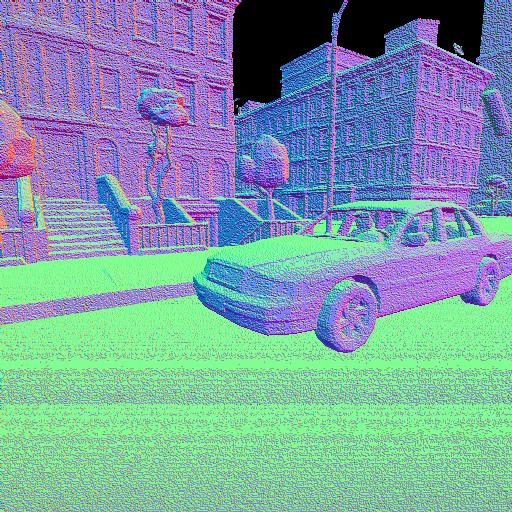} & 
        \includegraphics[width=2.3cm]{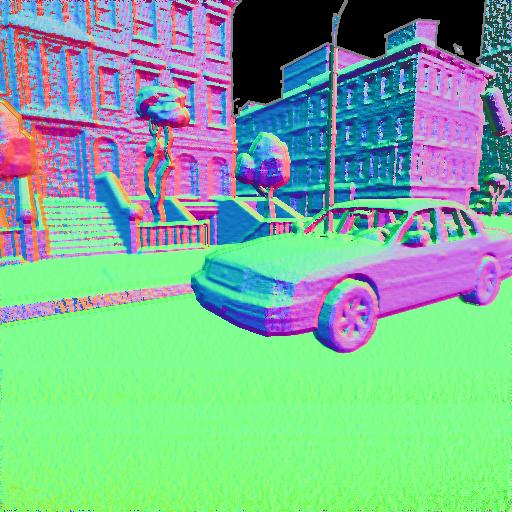} \\
        \includegraphics[width=2.3cm]{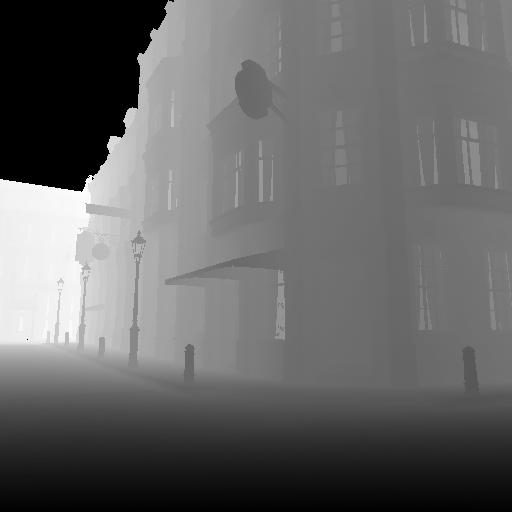} &
        \includegraphics[width=2.3cm]{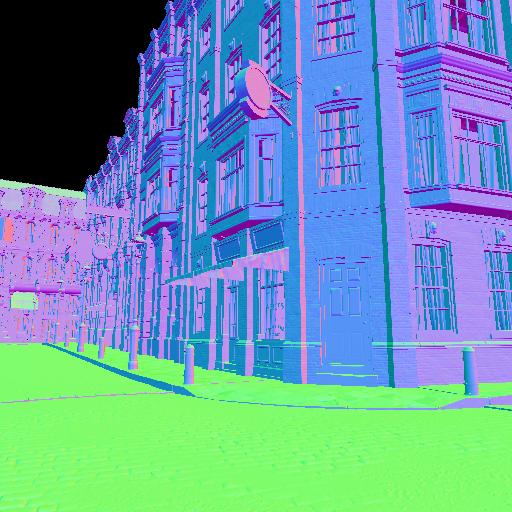} &
        \includegraphics[width=2.3cm]{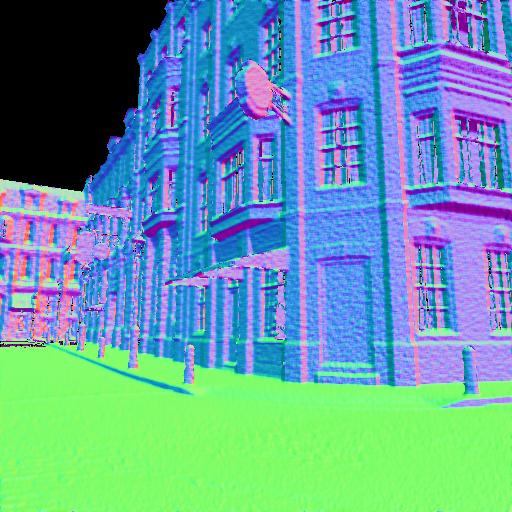} & 
        \includegraphics[width=2.3cm]{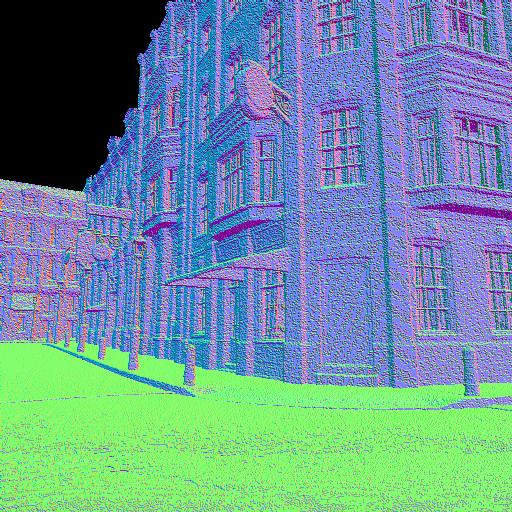} & 
        \includegraphics[width=2.3cm]{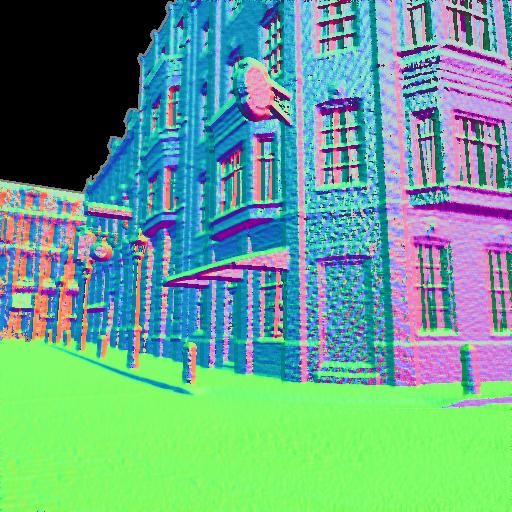} \\
        \includegraphics[width=2.3cm]{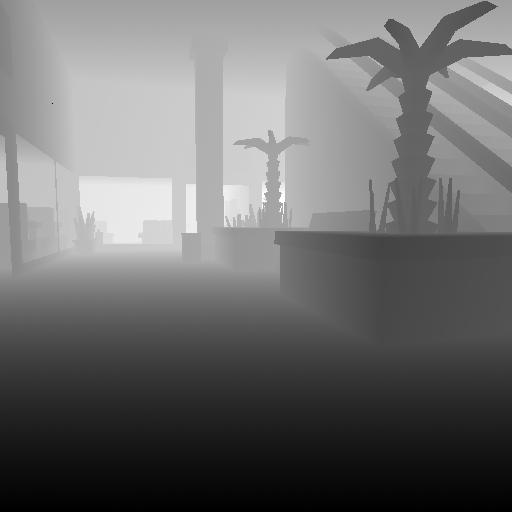} &
        \includegraphics[width=2.3cm]{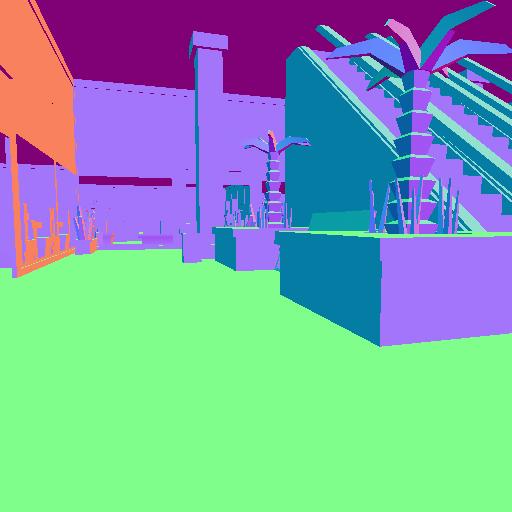} &
        \includegraphics[width=2.3cm]{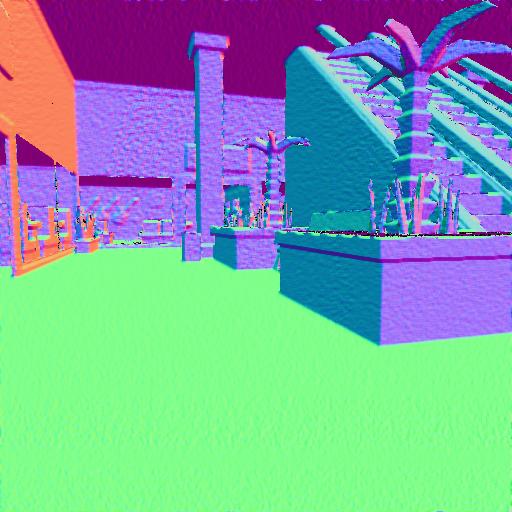} & 
        \includegraphics[width=2.3cm]{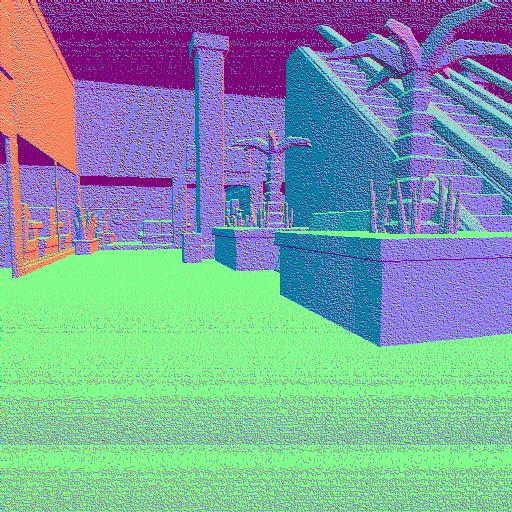} & 
        \includegraphics[width=2.3cm]{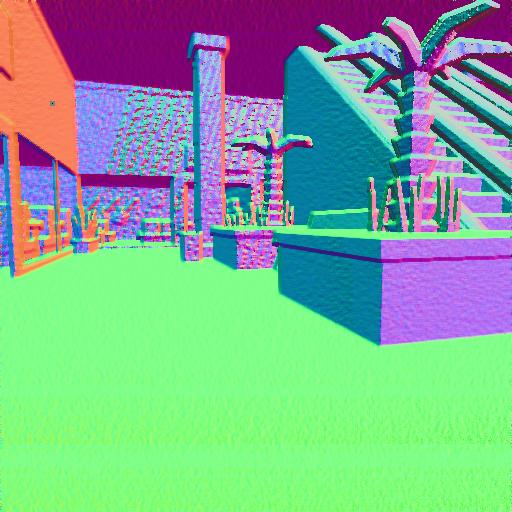} \\
        \includegraphics[width=2.3cm]{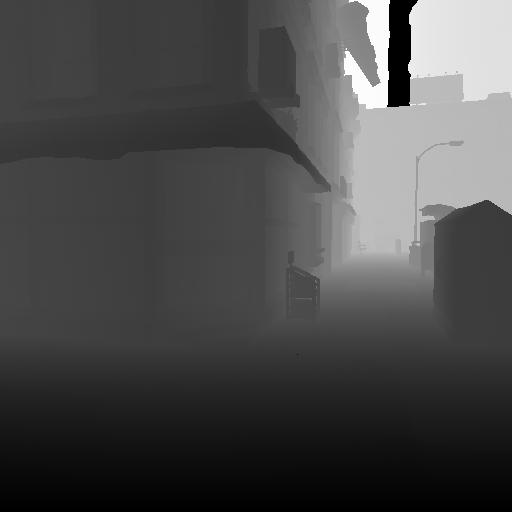} &
        \includegraphics[width=2.3cm]{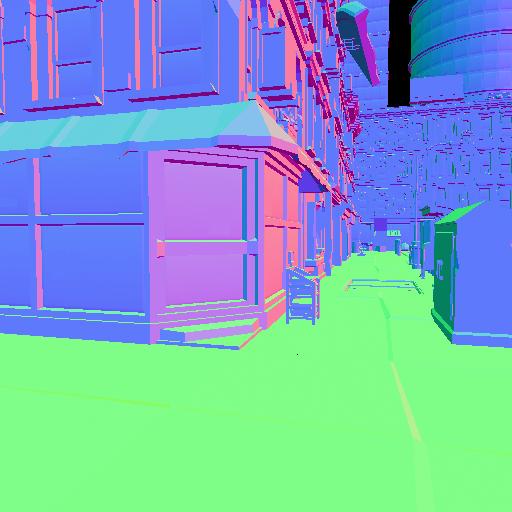} &
        \includegraphics[width=2.3cm]{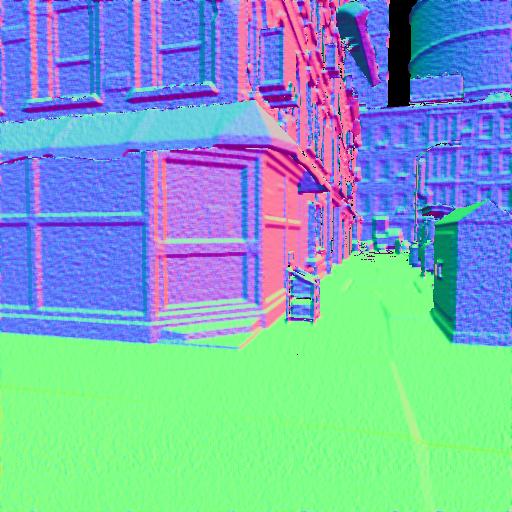} & 
        \includegraphics[width=2.3cm]{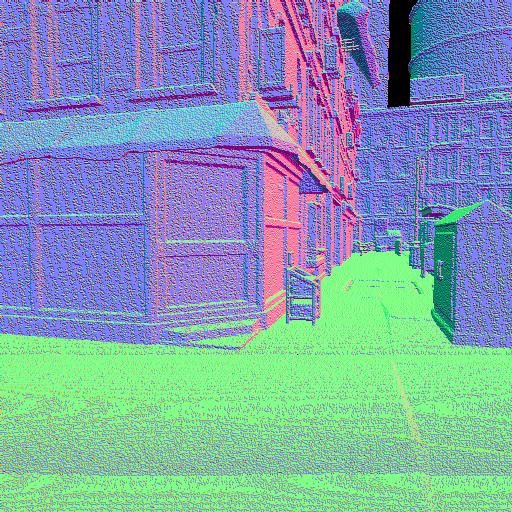} & 
        \includegraphics[width=2.3cm]{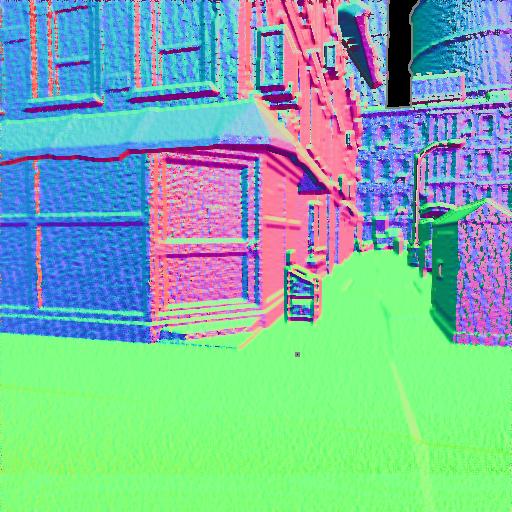} \\
    \end{tabular}}
    \caption{Normals-from-depth results on the 3D Ken Burns dataset \cite{Niklaus_TOG_2019}. Zoom for detail.}
    \label{fig:NfD}
\end{figure}

\paragraph{Normals-from-depth}
To evaluate normals-from-depth we use the synthetic data-set of Niklaus \etal \cite{Niklaus_TOG_2019}. This comprises realistic scene renderings and includes depth and normal maps. The normal maps are obtained by rasterising the ground truth mesh and so correctly handle depth discontinuities. The depth maps contain noise due to quantisation. We compute normals from this noisy depth and compare against the ground truth rasterised normal map. The combination of quantisation noise and depth discontinuities make the task surprisingly difficult.

\begin{wraptable}[10]{r}{65mm}
    \centering
    {\small
    \begin{tabular}{c|cc}
     & \multicolumn{2}{c}{\textbf{Scene}} \\
    {\bf Method} & city-walking & victorian-walking  \\
    \hline
    Ours &{\bf 15.73} &{\bf 19.84}\\ 
    FD (sc) &23.60 &26.38\\
    FD (fw)  &38.11 &39.86\\
    \cite{klasing2009comparison} &25.37 &30.06
    \end{tabular}}
    \caption{Median angular error of estimated surface normals on two scenes from 3D Ken Burns dataset \cite{Niklaus_TOG_2019}.}
    \label{tab:NfD}
\end{wraptable}

Qualitative results are shown in Figure \ref{fig:NfD}. Note that the finite difference (using forward difference) normal map is extremely noisy. Plane PCA \cite{klasing2009comparison} reduces noise but introduces planar discontinuities across depth boundaries while our result is smooth but preserves depth discontinuities. This is due to the 3D nearest neighbour filters. We show quantitative results for two scenes in Table \ref{tab:NfD}. Here we also include smoothed central finite difference. The proposed approach reduces error relative to the next best performing method by over 25\%.

\begin{figure}[!t]
    \centering
    \begin{tabular}{cccccc}
    \includegraphics[trim=80px 80px 80px 30px,clip=true,height=1.85cm]{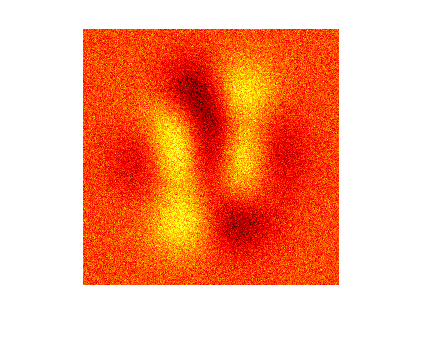} &
    \includegraphics[trim=80px 80px 80px 30px,clip=true,height=1.85cm]{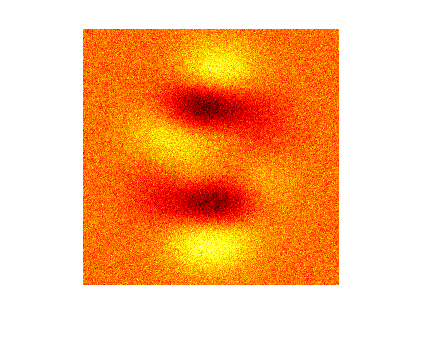} &
    \includegraphics[trim=80px 80px 80px 30px,clip=true,height=1.85cm]{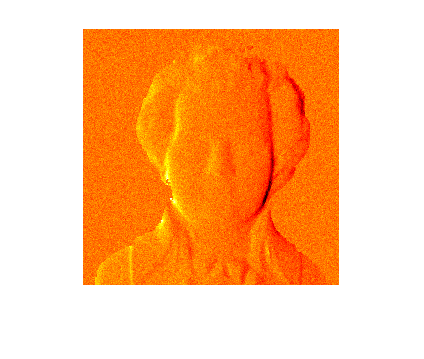} &
    \includegraphics[trim=80px 80px 80px 30px,clip=true,height=1.85cm]{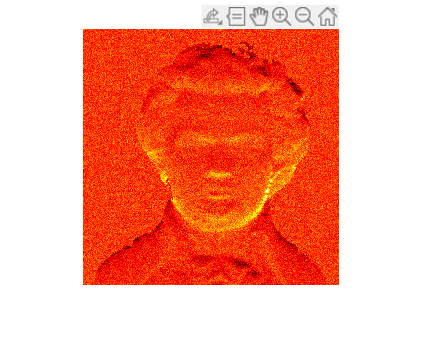} &
    \includegraphics[trim=80px 80px 80px 30px,clip=true,height=1.85cm]{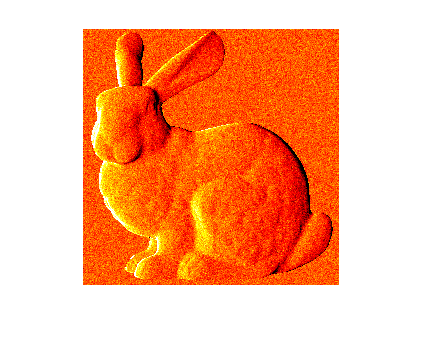} &
    \includegraphics[trim=80px 80px 80px 30px,clip=true,height=1.85cm]{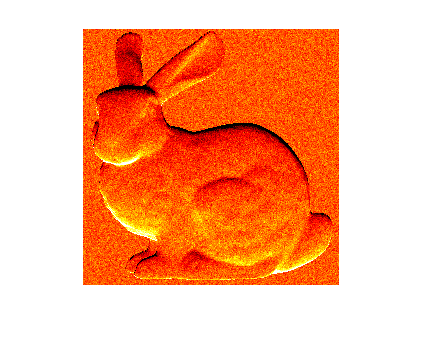} \\
    
    \includegraphics[trim=80px 80px 80px 30px,clip=true,height=1.85cm]{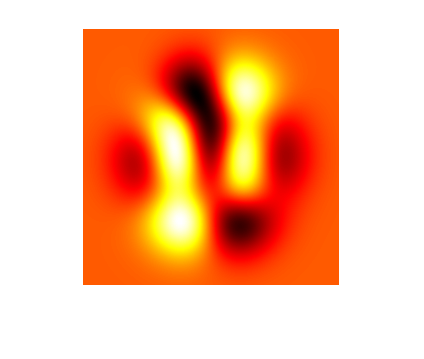} &
    \includegraphics[trim=80px 80px 80px 30px,clip=true,height=1.85cm]{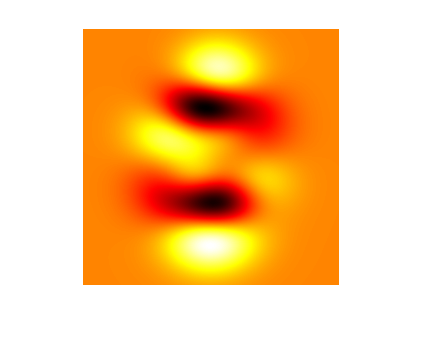} &
    \includegraphics[trim=80px 80px 80px 30px,clip=true,height=1.85cm]{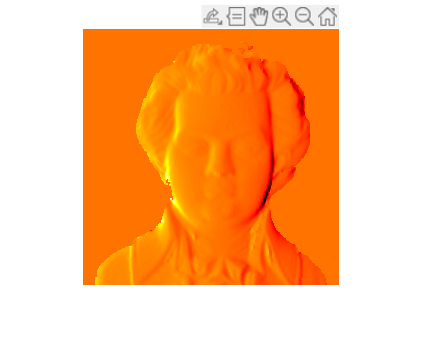} &
    \includegraphics[trim=80px 80px 80px 30px,clip=true,height=1.85cm]{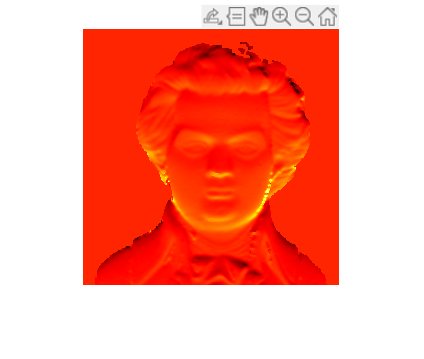} &
    \includegraphics[trim=80px 80px 80px 30px,clip=true,height=1.85cm]{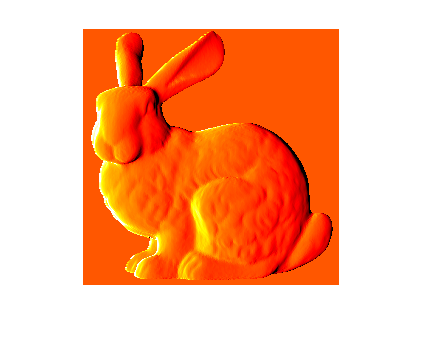} &
    \includegraphics[trim=80px 80px 80px 30px,clip=true,height=1.85cm]{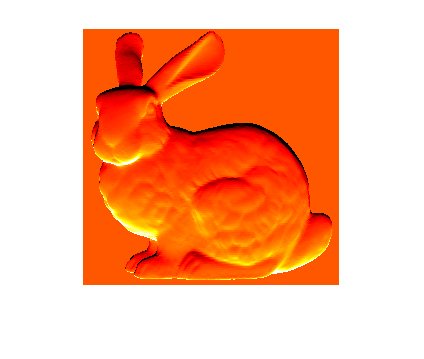}
    \end{tabular}
    \caption{First row presents the input noisy x and y gradient maps corresponding to the surfaces in Figure \ref{fig:syn_vis}. The second row shows the ground truth without noise.} 
    \label{fig:noisy_vis}
\end{figure}

\begin{figure}[!t]
    \centering
    \resizebox{\columnwidth}{!}{%
    \begin{tabular}{c@{\hspace{0.01\linewidth}}c@{\hspace{0.01\linewidth}}c@{\hspace{0.01\linewidth}}c@{\hspace{0.01\linewidth}}c@{\hspace{0.01\linewidth}}}
    \includegraphics[trim=100px 70px 90px 90px,clip=true, height=.15\linewidth]{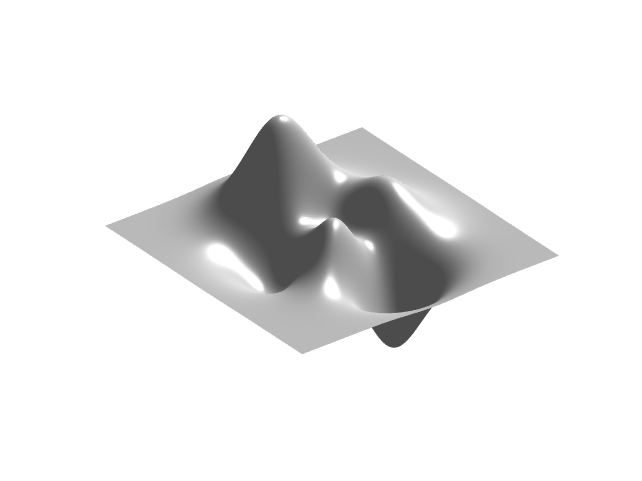} &
    \includegraphics[trim=100px 70px 90px 90px,clip=true, height=.15\linewidth]{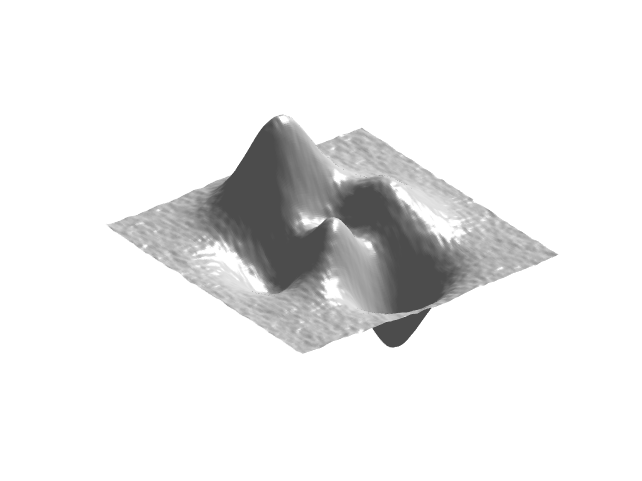} &
    \includegraphics[trim=100px 70px 90px 90px,clip=true, height=.15\linewidth]{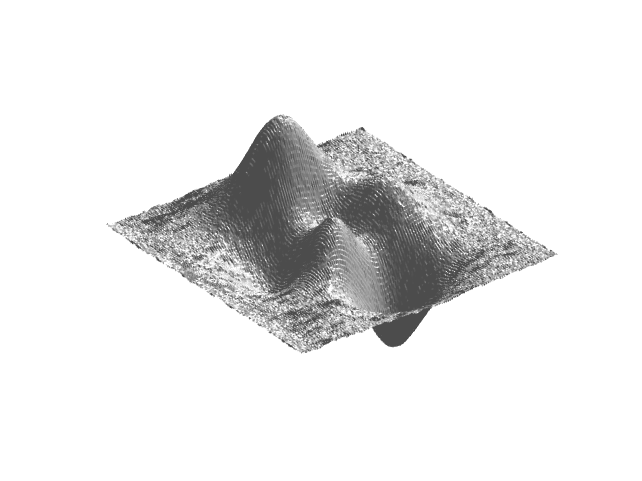} &
    \includegraphics[trim=100px 70px 90px 90px,clip=true, height=.15\linewidth]{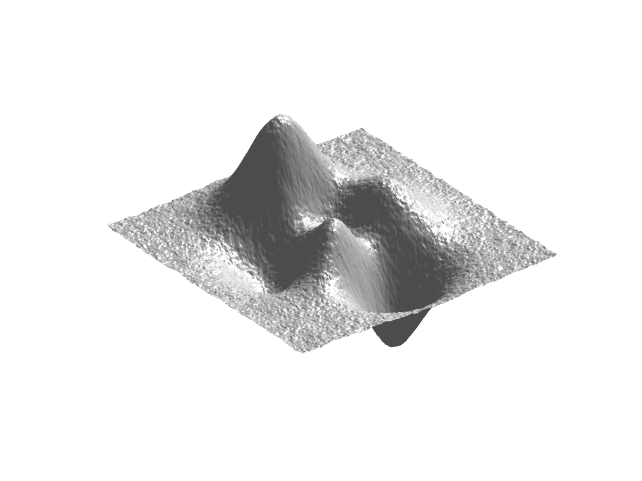} &
    \includegraphics[trim=100px 70px 90px 90px,clip=true, height=.15\linewidth]{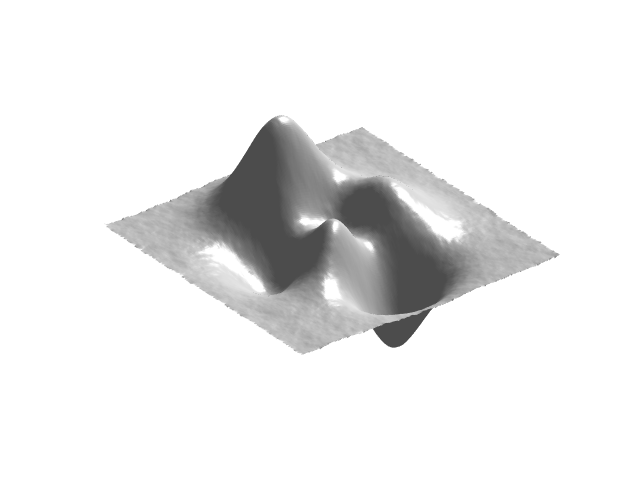} \\
    
    \includegraphics[trim=150px 70px 170px 90px,clip=true, height=.15\linewidth]{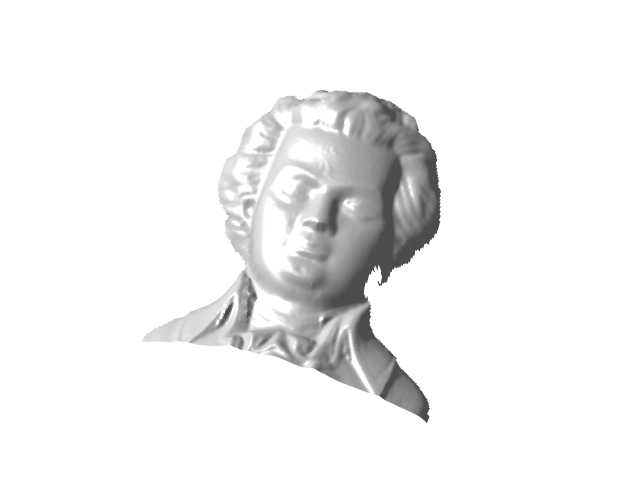} &
    \includegraphics[trim=150px 70px 170px 90px,clip=true, height=.15\linewidth]{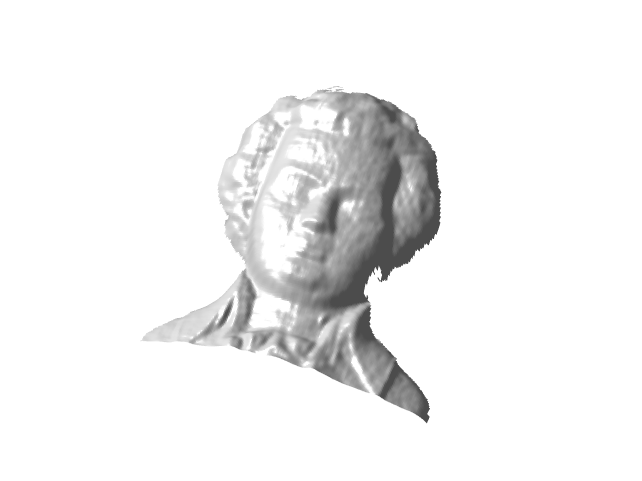} &
    \includegraphics[trim=150px 70px 170px 90px,clip=true, height=.15\linewidth]{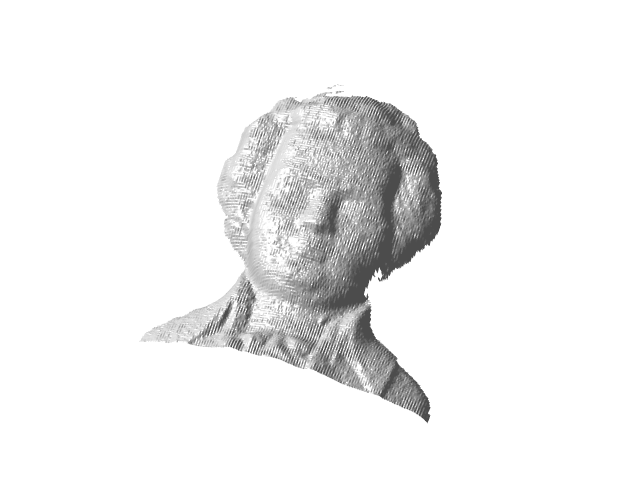} &
    \includegraphics[trim=150px 70px 170px 90px,clip=true, height=.15\linewidth]{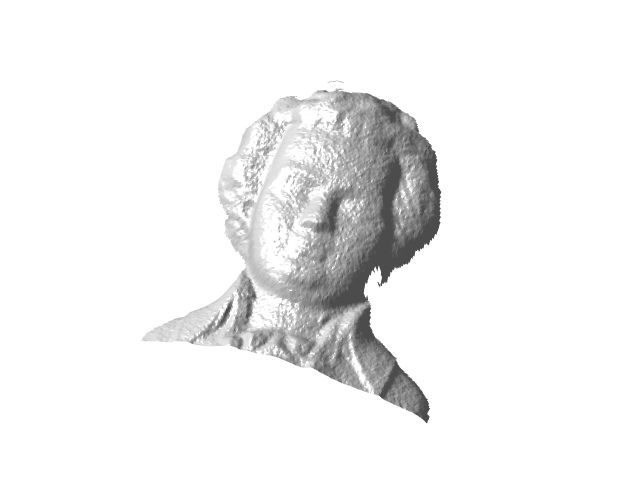} &
    \includegraphics[trim=150px 70px 170px 90px,clip=true, height=.15\linewidth]{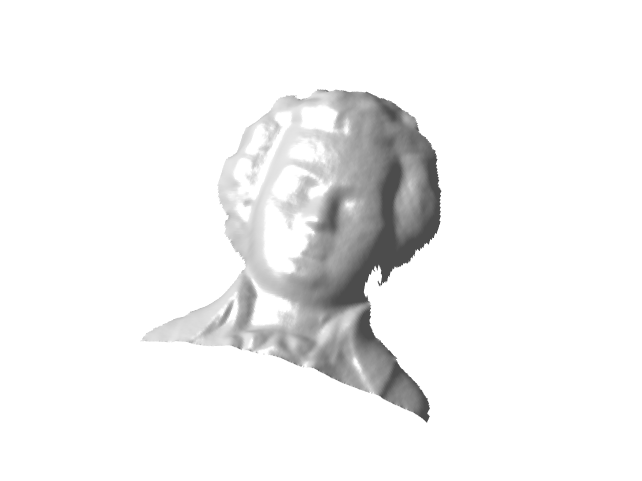} \\
    
    \includegraphics[trim=160px 70px 170px 120px,clip=true, height=.15\linewidth]{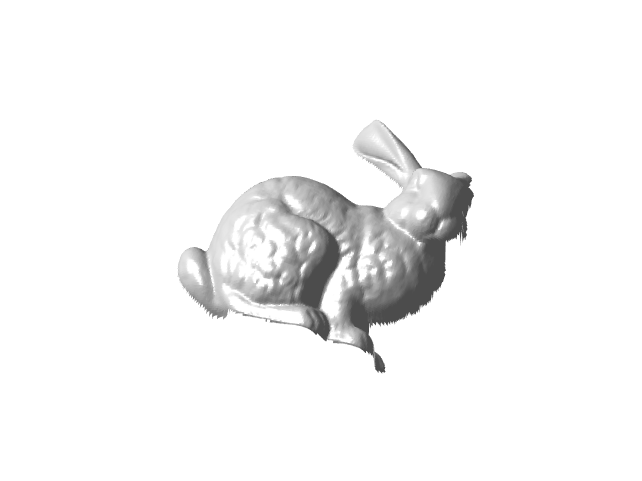} &
    \includegraphics[trim=160px 70px 170px 120px,clip=true, height=.15\linewidth]{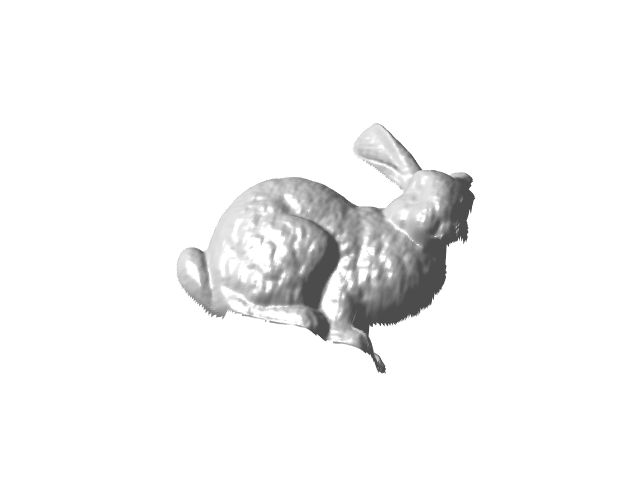} &
    \includegraphics[trim=160px 70px 170px 120px,clip=true, height=.15\linewidth]{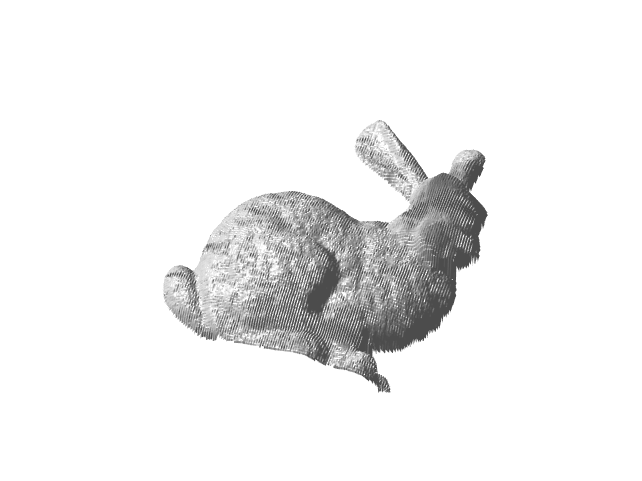} &
    \includegraphics[trim=160px 70px 170px 120px,clip=true, height=.15\linewidth]{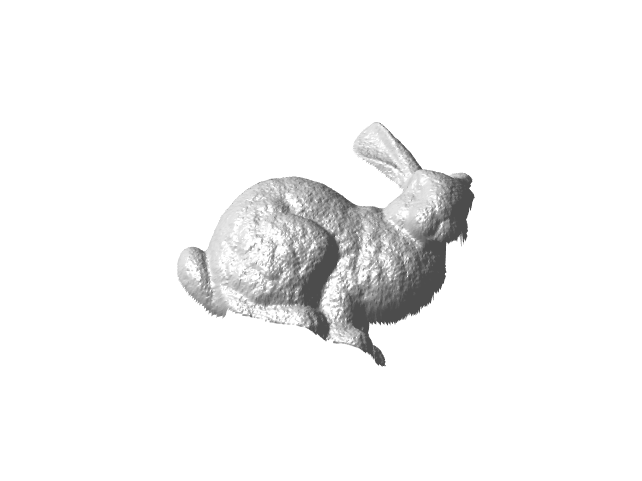} &
    \includegraphics[trim=160px 70px 170px 120px,clip=true, height=.15\linewidth]{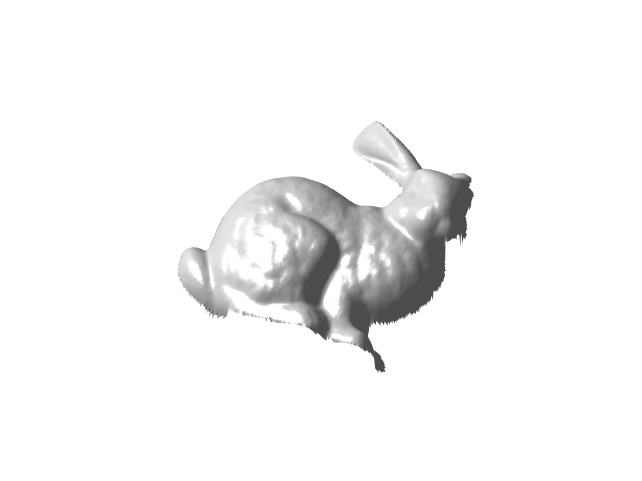} \\

    ground truth& Ours (SG) & \cite{harker2015regularized} &\cite{queau2018variational} & Ours (FD)
    \end{tabular}
    }
    \caption{Qualitative results on synthetic noisy data.} 
    \label{fig:syn_vis}
\end{figure}

\paragraph{Height-from-normals}
We evaluate our height-from-normals method on both synthetic and real data. We compare against two state-of-the-art methods \cite{harker2015regularized} and the best performing variant (Total Variation minimisation) of \cite{queau2018variational}. When applying these methods to perspective data we use the transformation proposed in \cite{queau2018normal} to reconstruct in the log domain before exponentiating to recover perspective depth. We compare against our approach using backward finite difference (FD) and the proposed Savotzky-Golay filters (SG).

\begin{wrapfigure}[17]{r}{65mm}
\vspace{-1cm}
    \centering
    \includegraphics[width=0.5\textwidth,clip=true]{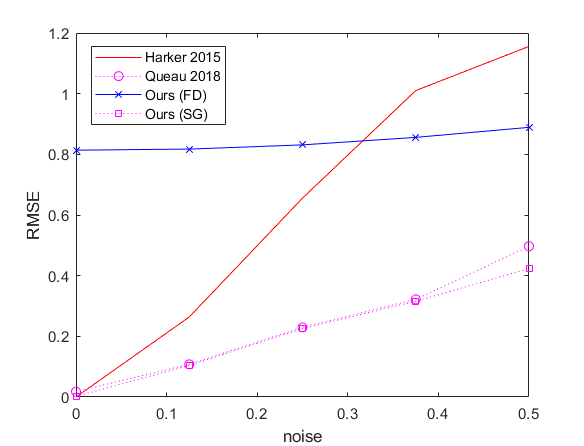}
    \caption{Root Mean Square Error between the recovered depth and the ground truth versus the standard deviation of the additive Gaussian noise. We use the synthetic image of the peaks with increasing Gaussian noise.}
    \label{fig:synthetic_RMSE}
\end{wrapfigure}

We begin by evaluating on synthetic data (peaks, Mozart and Stanford bunny) with Gaussian noise added to the input surface gradients (see Figure \ref{fig:syn_vis}). We show qualitative results in Figure \ref{fig:syn_vis}. Note that \cite{harker2015regularized} introduces a checkerboard effect in the presence of noise. This is due to the independence of adjacent pixel height estimates caused by approximating derivatives only along a single row or column. \cite{queau2018variational} is noisy due to having no explicit smoothness prior. Finite difference is smooth due to the strong Laplacian filter regularisation but this causes the surface to flatten. Our result preserves the global shape while also retaining local smoothness. We show the influence of varying the standard deviation of the noise in Figure \ref{fig:synthetic_RMSE}. \cite{harker2015regularized} degrades very quickly with noise. We slightly outperform \cite{queau2018variational} and note that our method is much more straightforward, requiring only the solution of a sparse linear system.

\begin{figure}[!t]
    \centering
    \resizebox{\columnwidth}{!}{%
    \begin{tabular}{c@{\hspace{0.01\linewidth}}c@{\hspace{0.01\linewidth}}c@{\hspace{0.01\linewidth}}}
    \includegraphics[height=.15\linewidth]{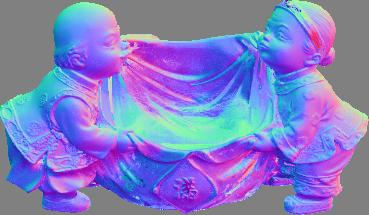} &
    \includegraphics[trim=120px 80px 80px 130px,clip=true,height=.15\linewidth]{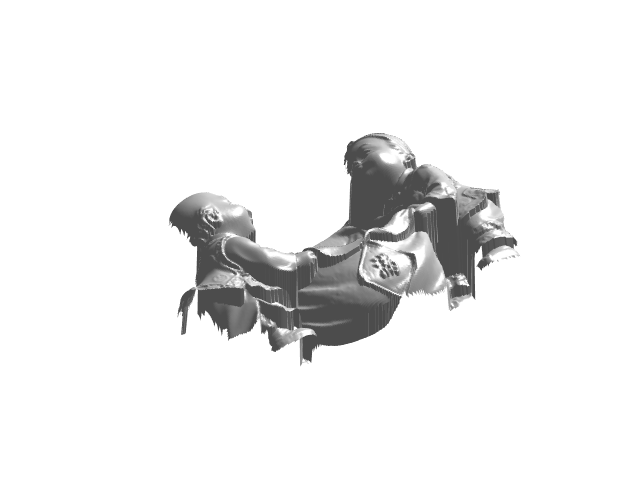} &
    \includegraphics[trim=120px 80px 80px 130px,clip=true,height=.15\linewidth]{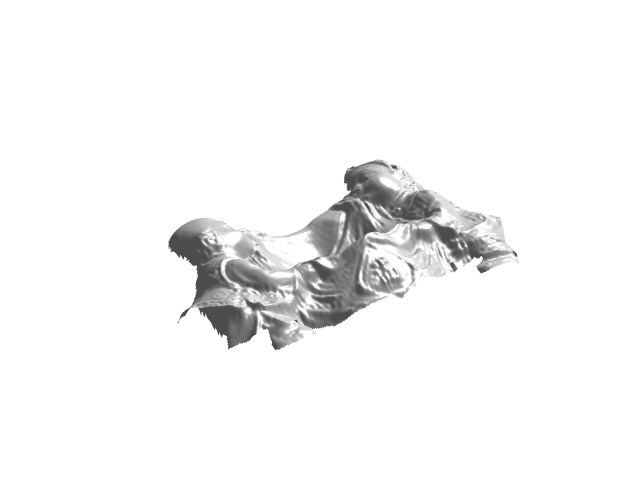} \\
    Input normals from \cite{ikehata2014photometric} & ground truth & ours (SG) \\
    \includegraphics[trim=120px 80px 80px 130px,clip=true,height=.15\linewidth]{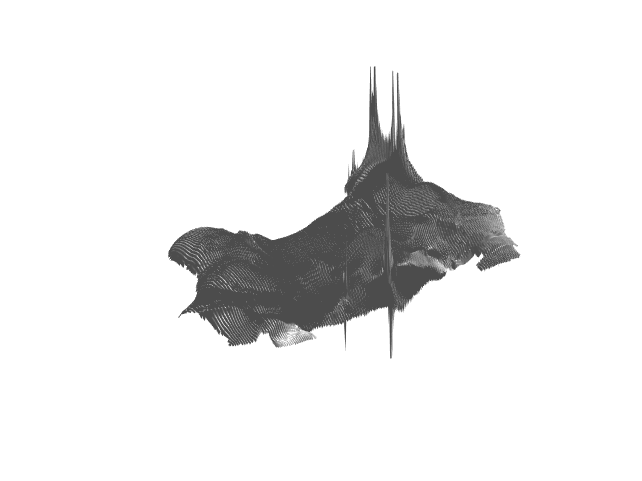} &
    \includegraphics[trim=120px 80px 80px 130px,clip=true,height=.15\linewidth]{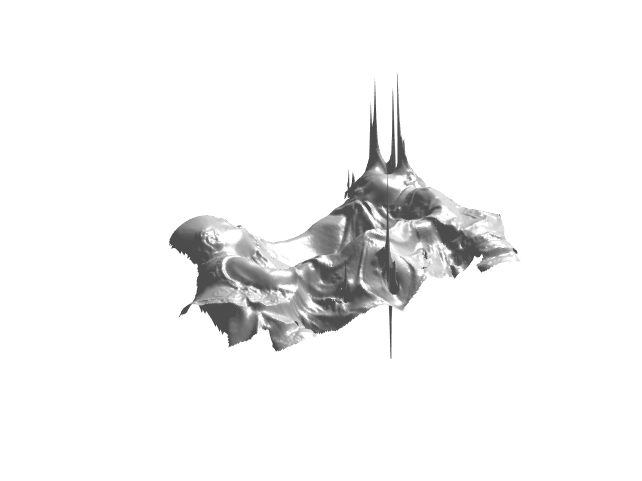} &
    \includegraphics[trim=120px 80px 80px 130px,clip=true,height=.15\linewidth]{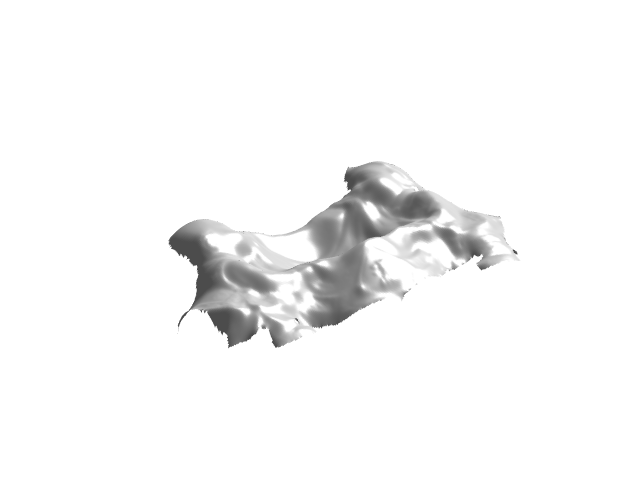} \\
    \cite{harker2015regularized} & \cite{queau2018variational} & ours (FD) \\
    \hline \\ [-1.5ex]
    \includegraphics[height=.15\linewidth]{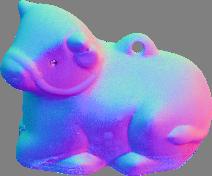} &
    \includegraphics[trim=120px 80px 80px 130px,clip=true,height=.15\linewidth]{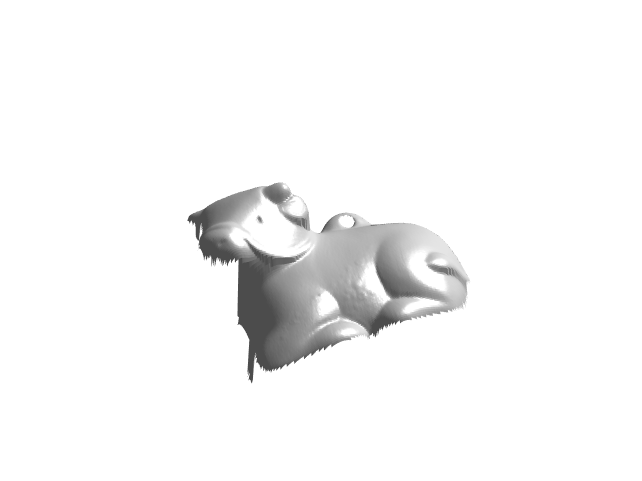} &
    \includegraphics[trim=120px 80px 80px 130px,clip=true,height=.15\linewidth]{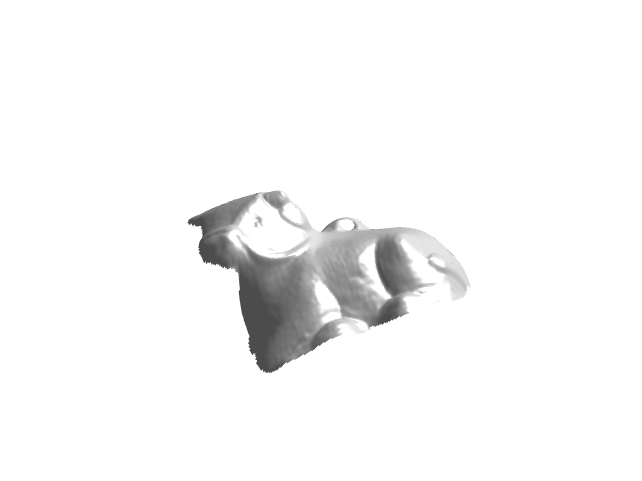} \\
    Input normals from \cite{ikehata2014photometric} & ground truth & ours (SG) \\
    \includegraphics[trim=120px 80px 80px 130px,clip=true,height=.15\linewidth]{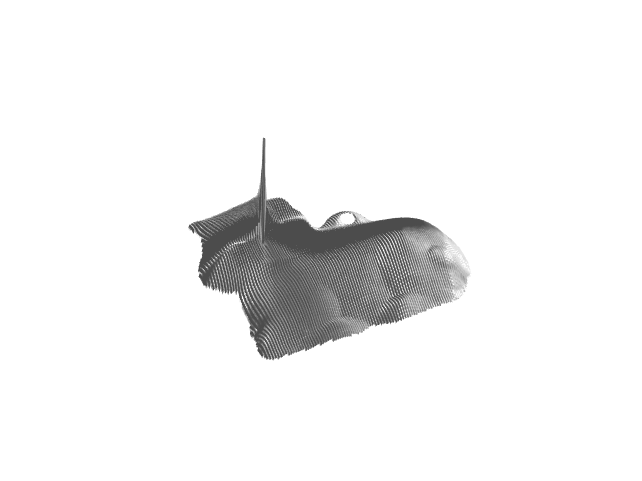} &
    \includegraphics[trim=120px 80px 80px 130px,clip=true,height=.15\linewidth]{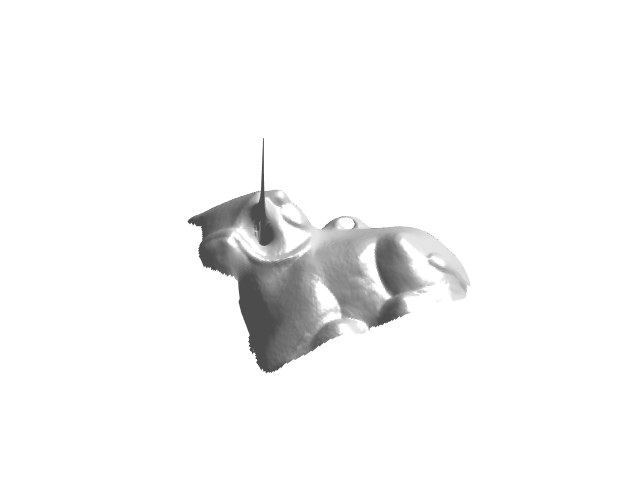} &
    \includegraphics[trim=120px 80px 80px 130px,clip=true,height=.15\linewidth]{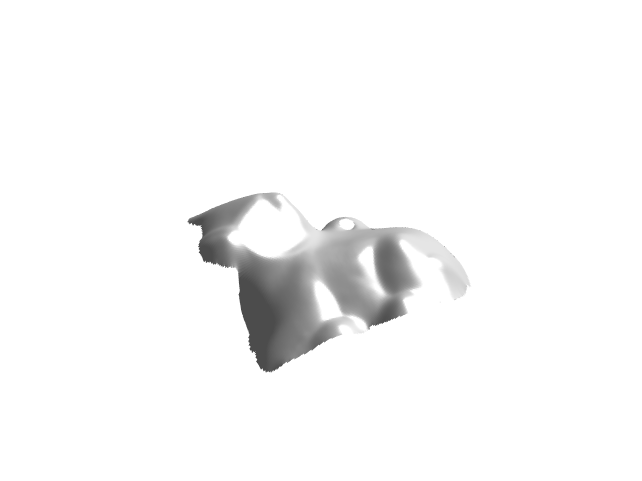} \\
    \cite{harker2015regularized} & \cite{queau2018variational} & ours (FD) \\
    \end{tabular}
    }
    \caption{Qualitative perspective height-from-normals results on real data.} 
    \label{fig:real_vis}
\end{figure}

Next, we evaluate on real data. In this case, it is a perspective reconstruction task. We use the surface normals estimated by \cite{ikehata2014photometric} on the DiLiGenT benchmark \cite{shi2019benchmark}. We use the camera parameters, foreground masks and ground truth depth provided with the benchmark. We integrate the estimated normals and compare the resulting depth to ground truth. Since the scale of the depth is unknown, we first compute the optimal scale between reconstruction and ground truth prior to computing the RMS error. We show qualitative results in Figure \ref{fig:real_vis}. Note that the nonlinear transformation causes spikes in the comparison methods. Our approach yields the visually best results, apart from smoothing across the discontinuities in the first example. We show quantitative results in Table \ref{tab:photoste}. Note that, in the cases where we are outperformed by \cite{queau2018variational}, their result often contains severe spike artefacts that create a poor visual reconstruction.

\begin{table}[t]
    \centering
    \begin{tabular}{cc|ccccccccc}
    &{\bf Method} & {\bf bear} & {\bf buddha} & {\bf cat} & {\bf cow} & {\bf goblet} & {\bf harvest} & {\bf pot 1} & {\bf pot 2} & {\bf reading} \\
    \cline{2-11}
    \multirow{3}{*}{\rotatebox[origin=c]{90}{Depth}} &
    \cite{harker2015regularized} &6.11 &6.26 &7.29 &4.82 &19.03 &11.72 &3.51 &2.69 &11.68 \\
    &\cite{queau2018variational} &5.55       &4.35       &5.90       &2.49       &15.49       &{\bf 11.45}&{\bf 2.78} &{\bf 1.75} &{\bf 9.66}\\
    &Ours (SG)                    &{\bf 5.42} &{\bf 3.97} &{\bf 5.53} &{\bf 2.13} &{\bf 14.90} &11.55     &3.07       &1.93 &10.29\\
    \cline{2-11}
    \multirow{3}{*}{\rotatebox[origin=c]{90}{Normal}} & \cite{harker2015regularized}&20.30      &24.06      &19.27       &21.10       &24.09 &55.50       &20.88       &20.42        &39.00 \\
    &\cite{queau2018variational} &20.06       &22.64       &18.60       &15.20       &19.84       &29.64       &20.26      &19.34        &24.18\\
    &Ours (SG)                    &{\bf 19.97}  &{\bf 22.02} &{\bf 18.38} &{\bf 14.78} &{\bf 19.21}      &{\bf 26.22} &{\bf 20.12} &{\bf 19.17}  &{\bf 23.57}\\
    \end{tabular}
    \caption{Perspective surface integration errors for depth (Root Mean Square Error in millimetres) and surface normal (median angular error in degrees) of recovered depth and surface normals of recovered depth relative to ground truth.}
    \label{tab:photoste}
\end{table}

\paragraph{Photometric stereo} Finally, we evaluate the effect of using our proposed filters in a photometric stereo experiment. We choose the photometric stereo method of Smith and Fang \cite{smith2016height}. This is not state-of-the-art (using only the Lambertian reflectance model) but it solves directly for an orthographic height map from a set of calibrated photometric stereo images and makes use of derivative matrices in this solution. Hence, it makes a good test case for our alternative derivative matrices. It works by taking ratios between pairs of intensity observations yielding linear equations in the surface gradient. Then, substituting numerical derivative approximations, solves for the least squares optimal height map. We use the authors original implementation which uses smoothed central difference derivatives and Laplacian smoothing filter. We compare this implementation with one in which the only modification we make is to replace the derivative and smoothing matrices with our K-nearest pixel, 2D Savitzky-Golay filters. We keep all other parameters fixed. We again use the DiLiGenT dataset \cite{shi2019benchmark}, this time running the photometric stereo algorithm on the input images, computing an orthographic height map, computing normals from this and calculating the mean angular error to ground truth. We show quantitative results in Figure \ref{tab:photoste}. Simply replacing the derivative and smoothing matrices significantly reduces error, sometimes by over 50\%.

\begin{table}[t]
    \centering
    \begin{tabular}{c|ccccccccc}
    {\bf Method} & {\bf bear} & {\bf buddha} & {\bf cat} & {\bf cow} & {\bf goblet} & {\bf harvest} & {\bf pot 1} & {\bf pot 2} & {\bf reading} \\
    \hline
    \cite{smith2016height} & 12.7 & 26.3 & 15.1 & 25.4 & 20.0 & 33.2 & 20.0 & 24.2 & 25.5 \\
    \cite{smith2016height} + SG filters & 7.33 & 14.8 & 8.43 & 23.7 & 16.4 & 29.2 & 9.88 & 14.7 & 14.0 \\
    \end{tabular}
    \caption{Photometric stereo evaluation. We compare the surface normal median angular error between the two methods and ground truth.}
    \label{tab:photoste}
\end{table}

\section{Conclusions}

In this paper we have explored alternatives to the widely used numerical derivative approximations. This often overlooked choice turns out to be significant in the performance of normals-from-depth, height-from-normals and shape-from-x. While we propose specific methods for these problems, the main takeaway from this paper is that any algorithm that uses sparse derivative matrices could plug in our matrices based on 2D Savitzky-Golay filters and see benefit. There are many possible extensions. Our approach does not consider or deal with discontinuities (apart from the 3D nearest neighbour extension for normals-from-depth). A hybrid between our approach and the discontinuity aware approach of \cite{queau2018variational} may be possible. Another interesting avenue is integrating better differentiation kernels into deep learning frameworks. Where a segmentation mask is available or estimated, it should be possible to apply appropriate kernels to each segment avoiding smoothing across depth discontinuities. The challenge here would be making it sufficiently efficient for use in deep learning as well as making the kernel selection differentiable. Finally, it would be interesting to see whether normals-from-depth can be learnt as a black box process and whether this outperforms our handcrafted kernels. 

\subsection*{Acknowledgements}

W. Smith is supported by a Royal Academy of Engineering/The Leverhulme Trust Senior Research Fellowship.

%
%
\bibliographystyle{splncs04}
\bibliography{refs}
\end{document}